\documentclass{article}
\PassOptionsToPackage{numbers, compress}{natbib}
\usepackage[preprint]{neurips_2026}

\usepackage[utf8]{inputenc}
\usepackage[T1]{fontenc}
\usepackage{hyperref}
\usepackage{url}
\usepackage{amsmath,amssymb,amsthm}
\usepackage{mathtools}
\usepackage{bm}
\usepackage{booktabs}
\usepackage{multirow}
\usepackage{graphicx}
\usepackage{subcaption}
\usepackage{algorithm}
\usepackage{algpseudocode}
\usepackage{enumitem}
\usepackage{pifont}
\usepackage{marvosym}
\usepackage{xcolor}
\usepackage{float}
\usepackage{wrapfig}
\usepackage{colortbl}
\usepackage{makecell}
\usepackage{placeins}
\usepackage{afterpage}
\usepackage{pgfplots}
\usepackage[most]{tcolorbox}
\pgfplotsset{compat=1.18}
\tcbuselibrary{listings,breakable}

\newcommand{\method}{SpatialFlow-GRPO}
\newcommand{\rewardmodel}{SFReward}
\newcommand{\rewarddata}{SFReward-14K}

\newcommand{\cmark}{\ding{51}}
\newcommand{\xmark}{\ding{55}}

\definecolor{bestcolor}{HTML}{E8F5E9}
\definecolor{avgcolor}{HTML}{E8F5E9}
\definecolor{codeback}{HTML}{F8F9FA}
\definecolor{codeframe}{HTML}{D0D7DE}
\definecolor{codetitle}{HTML}{57606A}
\newtcblisting{jsonbox}{%
  listing only,
  breakable,
  width=\linewidth,
  colback=codeback,
  colframe=codeframe,
  boxrule=0.4pt,
  arc=2pt,
  left=5pt,
  right=5pt,
  top=4pt,
  bottom=4pt,
  listing options={
    basicstyle=\ttfamily\footnotesize,
    columns=fullflexible,
    keepspaces=true,
    breaklines=true,
    showstringspaces=false
  }
}
\newtcblisting{compactjsonbox}{%
  listing only,
  width=\linewidth,
  colback=codeback,
  colframe=codeframe,
  boxrule=0.35pt,
  arc=3pt,
  left=6pt,
  right=6pt,
  top=4pt,
  bottom=4pt,
  before skip=3pt,
  after skip=3pt,
  listing options={
    basicstyle=\ttfamily\scriptsize,
    columns=fullflexible,
    keepspaces=true,
    breaklines=true,
    showstringspaces=false
  }
}
\newenvironment{promptblock}{%
  \par\smallskip
  \begin{tcolorbox}[
    breakable,
    colback=codeback,
    colframe=codeframe,
    boxrule=0.35pt,
    arc=3pt,
    left=6pt,
    right=6pt,
    top=5pt,
    bottom=5pt
  ]
  \footnotesize\ttfamily\raggedright
}{%
  \end{tcolorbox}
  \smallskip
}

\title{\textbf{\method{}: Where Spatial Credit Drives Image Editing}}

\makeatletter
\newcommand{\blfootnotetext}[1]{%
  \begingroup
    \renewcommand{\thefootnote}{}%
    \renewcommand{\@makefnmark}{}%
    \long\def\@makefntext##1{\noindent##1}%
    \footnotetext{#1}%
  \endgroup
}
\renewcommand{\@noticestring}{}
\makeatother

\author{%
  \begin{minipage}[t]{\textwidth}\centering
    Yankai Yang\textsuperscript{*\,1,2} \quad
    Yancheng Long\textsuperscript{*\,1,2} \quad
    Wei Chen\textsuperscript{2} \quad
    Xingyu Lu\textsuperscript{2} \quad
    Hongyang Wei\textsuperscript{2} \\[3pt]
    Bin Wen\textsuperscript{\textdagger\,\Letter\,2} \quad
    Fan Yang\textsuperscript{2} \quad
    Tingting Gao\textsuperscript{2} \quad
    Han Li\textsuperscript{2} \quad
    Shuo Yang\textsuperscript{\Letter\,1}
  \end{minipage}%
}

\begin{document}
\maketitle
\blfootnotetext{%
  \textsuperscript{*}Equal contribution.\quad
  \textsuperscript{\textdagger}Project leader.\quad
  \textsuperscript{\Letter}Corresponding authors.\\[1pt]
  \textsuperscript{1}Harbin Institute of Technology, Shenzhen.\quad
  \textsuperscript{2}Kuaishou Technology.\\[1pt]
  Correspondence to: Shuo Yang $<$shuoyang@hit.edu.cn$>$, Bin Wen $<$wenbin@kuaishou.com$>$.%
}

\begin{abstract}
Recent online reinforcement learning has substantially improved image editing quality. However, existing Flow-GRPO-style methods usually rely on a single whole-image reward, which makes fine-grained editing optimization difficult. We observe that a key obstacle in image editing is this spatial uniformity assumption: a whole-image reward cannot distinguish how different spatial regions contribute to image quality. To address this issue, we propose \method{}, a training framework that introduces spatially fine-grained reward feedback. The framework converts region-aware rewards into semantic-region-level optimization signals and aligns region advantages with the corresponding latent positions during policy updates. We also train a region-aware reward model, \rewardmodel{}, construct \rewarddata{} with region-annotated editing samples, and introduce MultiEditBench to evaluate multi-region editing ability. On OmniGen2 and FLUX.2-klein-4B, \method{} outperforms Flow-GRPO on GEdit-Bench, ImgEdit-Bench, and MultiEditBench. The results show that \method{} converts local feedback into spatially aligned update signals and improves editing quality.
\end{abstract}

\section{Introduction}
\label{sec:intro}

Diffusion models have made rapid progress in image generation and editing~\citep{ho2020ddpm,song2021score,rombach2022ldm,esser2024flux}. Image editing has therefore become a core ability of visual generative models~\citep{brooks2023instructpix2pix,meng2022sdedit,couairon2022diffedit,hertz2023p2p,fu2024mgie}. Unlike open-ended text-to-image generation, editing quality is often spatially non-uniform. Different regions in the same output may show different degrees of completion, failure, or quality degradation. Thus, post-training for image editing needs more than a global preference signal. It also needs fine-grained feedback that can indicate quality differences across spatial regions.

Recent work has introduced online reinforcement learning (RL) into diffusion model post-training. Reward models provide preference, aesthetic, or editing-quality feedback to improve generation quality and instruction alignment~\citep{black2024ddpo,fan2024dpok,imagedoctor2025}. However, methods such as Flow-GRPO~\citep{flowgrpo2025} and DanceGRPO~\citep{dancegrpo2025} usually treat an edited image as a single outcome. They use one whole-image reward to compute a sample-level advantage and apply the same advantage to all latent positions. This design implies a \textbf{spatial uniformity assumption}: different regions in the image share the same credit signal. This assumption limits fine-grained optimization in editing tasks, because a single whole-image score cannot determine which target region accounts for editing success or failure, or whether the change reflects preservation quality in unedited regions.

\begin{figure}[t]
\centering
\includegraphics[width=\linewidth]{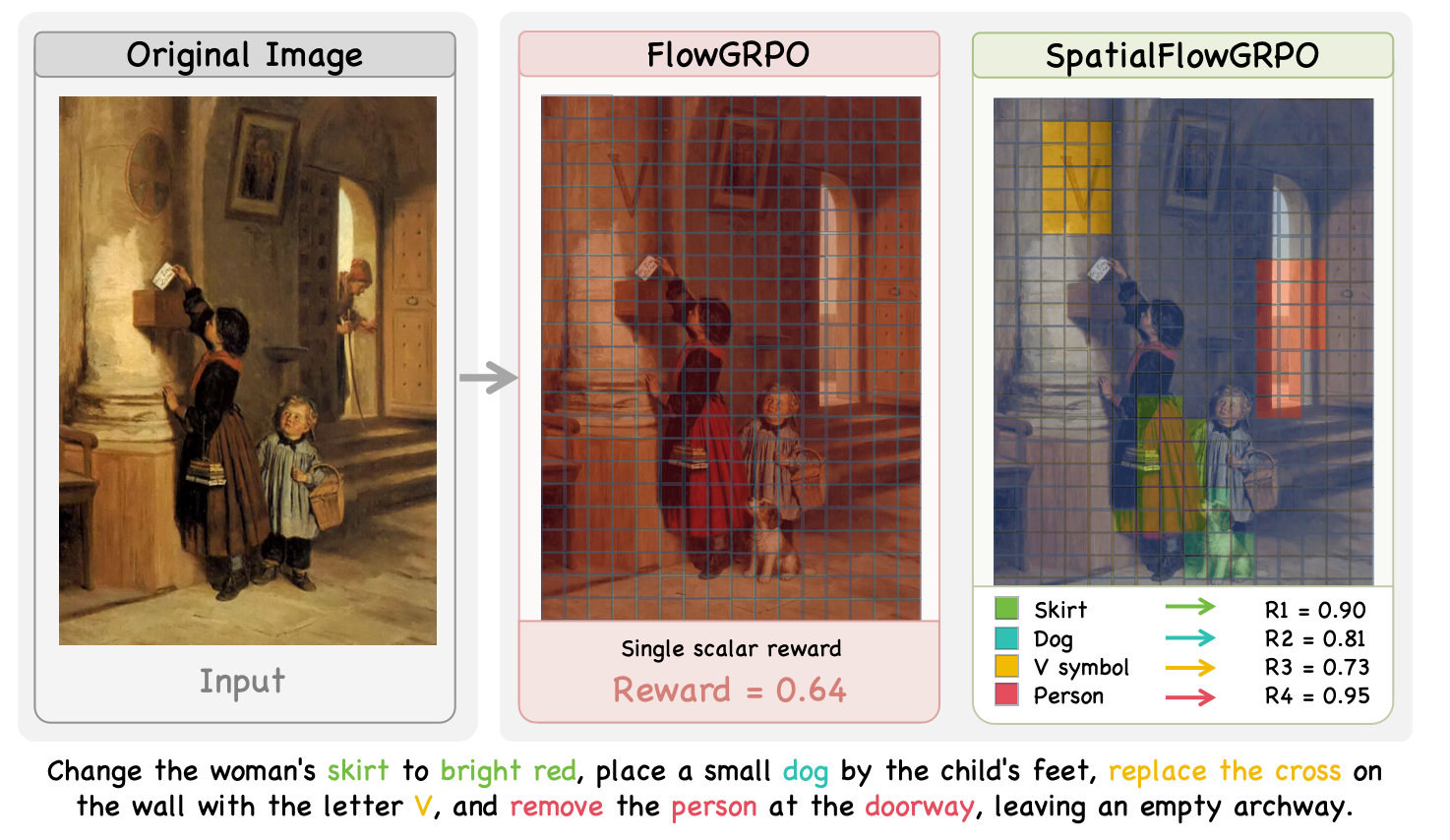}
\caption{\textbf{Motivation.} In multi-region editing, different regions may have different quality outcomes. Flow-GRPO collapses them into one scalar reward, while \method{} attaches reward feedback to semantic regions and enables spatially localized credit assignment.}
\label{fig:motivation}
\end{figure}

Figure~\ref{fig:motivation} shows a multi-region editing example: the instruction modifies the skirt, adds a dog, replaces a wall symbol with the letter V, and removes a person from the doorway. Flow-GRPO receives only one scalar reward for the whole result, so local successes and failures are collapsed into the same feedback. Region-level feedback can instead assign separate scores to the affected semantic regions, making the update signal spatially localizable. We call this the \textbf{spatial credit assignment dilemma}: quality differences in image editing have spatial structure, but the RL supervision signal and update rule lack matching spatial resolution. This problem is closely related to credit assignment and reward design in reinforcement learning~\citep{schulman2016gae}.

Based on this observation, we propose \method{}, which extends the sample-level relative optimization of Flow-GRPO to spatially structured region-level optimization. The framework uses \rewardmodel{} to produce quality feedback tied to semantic regions. It compares semantically corresponding regions among candidates sampled for the same instruction, producing localized region advantages. \method{} further preserves this spatial correspondence in the policy objective. Region advantages are applied to the corresponding latent positions, and region-consistent policy ratios with region-weighted aggregation reduce the dilution of local signals by whole-image averaging. As a result, RL post-training no longer updates the model only according to whole-image quality. It can use region-level quality differences for more precise spatial credit assignment.

The main contributions are summarized as follows:
\begin{enumerate}[leftmargin=18pt]
  \item \textbf{Identifying the spatial uniformity limitation.} We identify that whole-image rewards in Flow-GRPO-style methods force different spatial regions to share the same credit signal. This is a key factor that limits fine-grained optimization for image editing.

  \item \textbf{Region-level RL training framework.} We propose \method{}, which integrates region-aware rewards, semantic-group advantage estimation, and a latent-region-aligned policy objective into online RL post-training. This allows local feedback to serve as localized training signals for model updates.

  \item \textbf{Region reward model, data, and benchmark.} We train the region-aware reward model \rewardmodel{}, build \rewarddata{} for region reward learning, and introduce MultiEditBench for diagnosing fine-grained editing ability. On OmniGen2 and FLUX.2-klein-4B, \method{} outperforms Flow-GRPO on GEdit-Bench, ImgEdit-Bench, and MultiEditBench.
\end{enumerate}

\section{Related work}
\label{sec:related}

\paragraph{Image editing.}
Image editing modifies a source image according to text while preserving naturalness and semantic consistency. Prior methods use noise perturbation, masks, or attention control~\citep{meng2022sdedit,couairon2022diffedit,hertz2023p2p}, and instruction-based data and models further broaden editing ability~\citep{brooks2023instructpix2pix,zhang2023magicbrush,fu2024mgie}. Recent systems scale data, model interfaces, and task coverage~\citep{zhao2024ultraedit,huang2024smartedit,hui2024hqedit,omnigen2_2025,step1xedit2025,shi2024seededit,sheynin2024emuedit,wei2025omniedit}. We study how to post-train such editors with spatially fine-grained RL feedback.

\paragraph{RL post-training for diffusion models.}
RL post-training directly optimizes diffusion policies with reward or preference signals~\citep{black2024ddpo,fan2024dpok,wallace2024diffusiondpo}, including image-editing variants~\citep{li2024instructrl4pix,hpedit2026}. DanceGRPO~\citep{dancegrpo2025} and Flow-GRPO~\citep{flowgrpo2025} extend GRPO to visual generation and flow matching, while ImageDoctor~\citep{imagedoctor2025} studies diagnosis-oriented visual evaluation. These methods improve alignment but usually compress each edited image into one sample-level reward, which can hide local quality variation.

\paragraph{Fine-grained credit assignment and region-aware rewards.}
Fine-grained credit assignment improves RL efficiency and stability. GRPO-style reasoning systems show the value of relative optimization~\citep{shao2024deepseekmath,guo2025deepseekr1,dapo2025}, and B2-DiffuRL densifies sparse diffusion rewards over timesteps~\citep{b2diffurl2025}; however, image editing also requires spatial credit. Visual reward models have become stronger~\citep{xu2023imagereward,kirstain2023pickscore,wu2023hpsv2,editscore2025,editreward2025,jrm2026,spatialreward2026}, and SpatialReward already provides structured region scores. Our contribution is complementary: we train an editing-oriented region evaluator and convert its structured scores into comparable region advantages while preserving spatial correspondence during policy updates.

\paragraph{Evaluation benchmarks and fine-grained editing diagnosis.}
Editing evaluation is also moving from overall scores toward detailed diagnosis. Existing benchmarks cover real user requests, general editing, multi-attribute/object editing, data quality, automatic evaluation, interpretable scoring, and spatial preservation~\citep{step1xedit2025,imgedit2025,cai2024paralleledits,hui2024hqedit,ma2024i2ebench,viescore2024,spatialreward2026}. They measure overall quality well, but rarely isolate whether a model handles spatially distinct goals as editing complexity increases. We introduce MultiEditBench to evaluate this fine-grained ability under multi-region edits.

\begin{figure}[t]
\centering
\includegraphics[width=\linewidth]{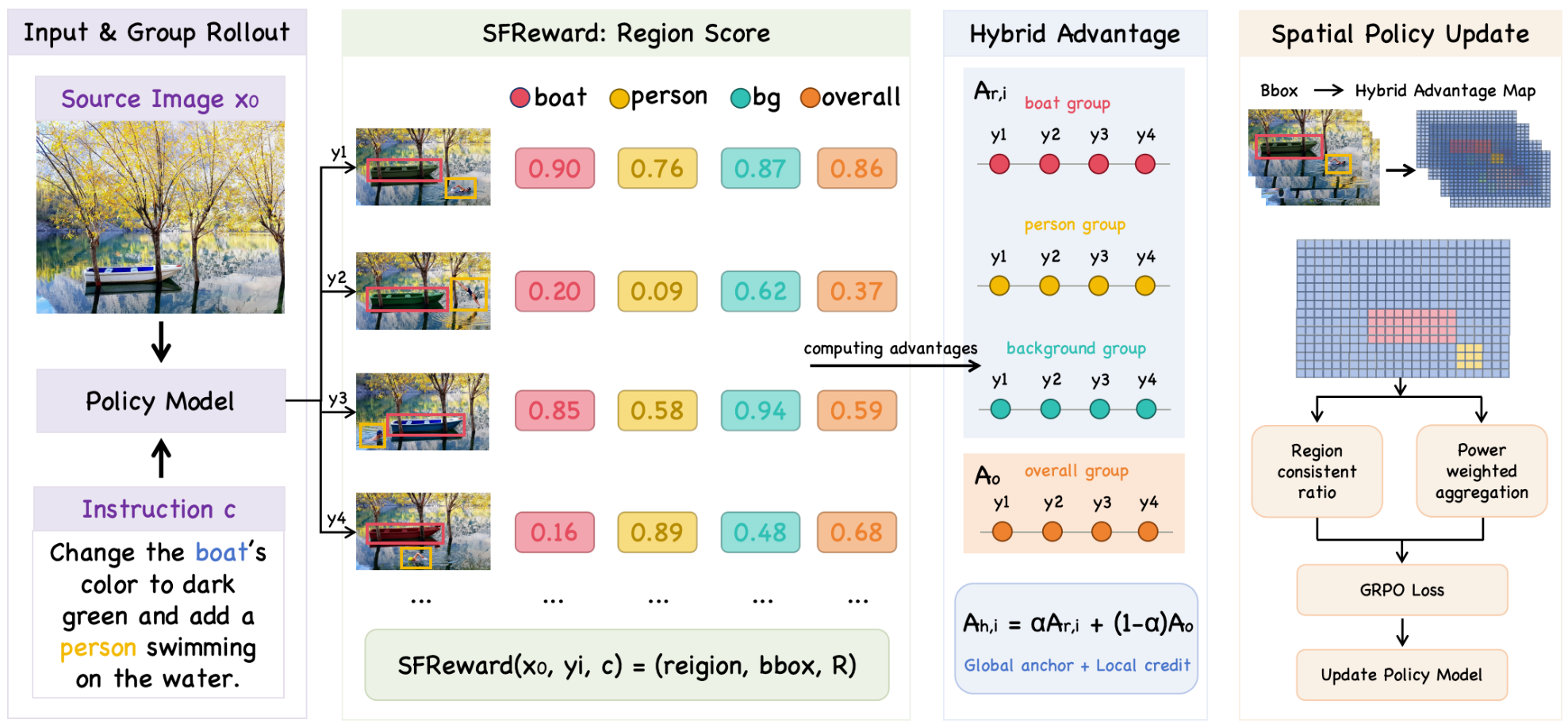}
\caption{\textbf{Overview of \method{}.} The policy samples a group of edited images for each instruction, and \rewardmodel{} returns region boxes, semantic labels, and scores. Region and global scores are converted into hybrid advantages, mapped back to latent regions, and optimized through region-consistent ratios and power-weighted aggregation.}
\label{fig:overview}
\end{figure}

\section{Method}
\label{sec:method}

The goal of \method{} is to turn whole-image relative optimization in GRPO into spatial credit assignment over semantic regions. Given an editing instruction, a source image, and $G$ candidate outputs sampled from the current policy, \rewardmodel{} predicts semantic regions to be scored, their bounding boxes, and rewards $\{R_{i,r}\}$ for each output. \method{} compares semantically corresponding regions under the same instruction and combines region-level advantages with a global quality anchor. The resulting hybrid advantages are aligned with the corresponding latent positions and used in the policy objective. In this way, the reward no longer affects the whole image only at the sample level. Instead, it can guide model updates at the region level. The overall framework is shown in Figure~\ref{fig:overview}.

\subsection{Problem setup and structured region reward}
\label{sec:prelim}
\label{sec:region_reward}

We follow the online training paradigm of Flow-GRPO~\citep{flowgrpo2025}. Given an editing instruction $c$ and a source image, the policy model generates $G$ edited images in each sampling round, together with their reverse diffusion trajectories $\{x_t^i\}_{t=0}^T$~\citep{ho2020ddpm,lipman2023flow}. Standard Flow-GRPO uses a sample-level reward $R_i$ to compute the within-group advantage:
\[
\hat{A}_i=\frac{R_i-\mu_R}{\sigma_R+\epsilon},
\]
and optimizes a PPO/GRPO-style clipped surrogate over all timesteps~\citep{schulman2017ppo}:
\begin{equation}
\label{eq:flowgrpo_base}
J_{\text{Flow-GRPO}}(\theta)=
\frac{1}{GT}\sum_{i=1}^{G}\sum_{t=0}^{T-1}
\min\!\left(
r_{i,t}(\theta)\hat{A}_i,\;
\operatorname{clip}(r_{i,t}(\theta),1-\epsilon_-,1+\epsilon_+)\hat{A}_i
\right)
-\lambda_{\mathrm{kl}}\mathcal{L}_{\mathrm{kl}},
\end{equation}
where
\[
r_{i,t}(\theta)=
\frac{p_\theta(x_{t-1}^i\!\mid x_t^i,c)}
{p_{\theta_{\text{old}}}(x_{t-1}^i\!\mid x_t^i,c)}.
\]
This objective fits global preference feedback but loses local editing success and failure in multi-region editing. Our Flow-GRPO baseline uses the sample-level score from the same \rewardmodel{} as $R_i$, while \method{} uses region scores to construct region-level advantages and objectives.

\begin{figure}[t]
\centering
\includegraphics[width=\textwidth]{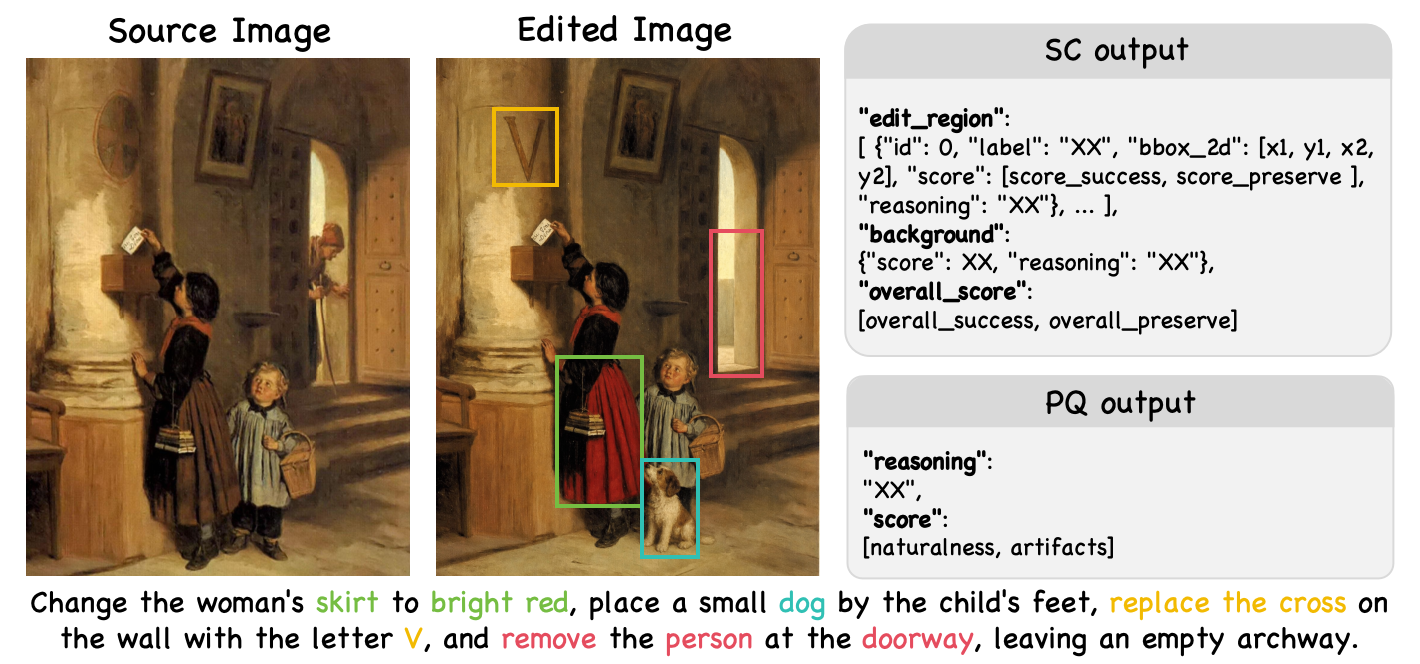}
\caption{\textbf{Structured output format of \rewardmodel{}.}}
\label{fig:reward_output_format}
\end{figure}

\rewardmodel{} outputs two types of structured scores for each edited result. Semantic consistency (SC) evaluates whether the edit succeeds and whether it avoids over-editing. Perceptual quality (PQ) evaluates image naturalness and artifacts. The output format is shown in Figure~\ref{fig:reward_output_format}.
Let $\mathbf{o}_i$ denote the \texttt{overall\_score} from SC, and let $\mathbf{q}_i$ denote the \texttt{score} from PQ. The Flow-GRPO baseline and the reward-model evaluation use a VIEScore-style aggregation~\citep{viescore2024} to combine global SC and PQ scores into a sample-level reward:
\begin{equation}
\label{eq:sample_reward}
R_i^{g}=
\frac{\sqrt{
\min(\mathbf{o}_i)\cdot \min(\mathbf{q}_i)
}}{C},
\end{equation}
where $C$ is a fixed normalization constant.

Unlike the sample-level reward in Equation~\eqref{eq:sample_reward}, \method{} computes a separate reward for each edited region and for the background. \rewardmodel{} provides semantic-consistency scores for the regions in \texttt{edit\_region}, a background-preservation score, and image-level perceptual-quality scores. For sample $i$, let $\mathcal{R}_i=\mathcal{R}_i^{\text{fg}}\cup\{\text{bg}\}$, let $\mathrm{IF}_{i,r}$ denote the semantic-consistency score of foreground region $r$, let $\mathrm{IF}_{i,\text{bg}}$ denote the background score, and let $\mathrm{AES}_i=\min(\mathbf{q}_i)$ denote the perceptual-quality term. The region reward is
\begin{equation}
\label{eq:region_reward}
R_{i,r}=
\frac{\sqrt{\phi(\mathrm{IF}_{i,r})\cdot\mathrm{AES}_i}}{C},
\qquad
\phi(\mathrm{IF}_{i,r})=
\begin{cases}
\min(\mathrm{IF}_{i,r}), & r\in\mathcal{R}_i^{\text{fg}},\\
\mathrm{IF}_{i,\text{bg}}, & r=\text{bg},
\end{cases}
\end{equation}
where $\phi$ selects the local semantic-consistency signal for each foreground region and the preservation signal for the background. $\mathrm{AES}_i$ is computed from image-level perceptual-quality scores, since naturalness, artifacts, and aesthetic quality are judged from the whole edited image rather than from isolated regions. It therefore serves as a global quality factor shared by all region rewards. \rewardmodel{} is trained on \rewarddata{}, a 14,276-sample region-annotated editing dataset; details are provided in Appendix~\ref{sec:reward_details}.

\subsection{Semantic-region advantage estimation}
\label{sec:region_adv}
\label{sec:hybrid_adv}

Rewards are comparable only within the same instruction and semantic label. \method{} therefore normalizes rewards by $(\text{instruction}, \text{label})$. Given instruction $p$ and label $l$, we define:
\begin{equation}
\label{eq:semantic_group}
\mathcal{S}_{p,l}=\{R_{i,r}\mid c_i=p,\; l_{i,r}=l\}.
\end{equation}
When $|\mathcal{S}_{p,l}|\ge2$, the region advantage is
\begin{equation}
\label{eq:region_adv}
\hat{A}_{i,r}=
\frac{R_{i,r}-\mu_{p,l}}{\sigma_{p,l}+\epsilon},
\end{equation}
where $\mu_{p,l}$ and $\sigma_{p,l}$ are the mean and standard deviation within the semantic group. Since grouping is within the same instruction, no predefined global semantic taxonomy is required.

Region advantages capture local quality differences, but pure local scores can weaken composition, identity preservation, and naturalness. We therefore mix each region advantage with the global advantage:
\begin{equation}
\label{eq:hybrid_adv}
\hat{A}_{i,r}^{\text{mix}}=(1-\alpha)\hat{A}_i^{g}+\alpha\hat{A}_{i,r},\quad \alpha\in[0,1].
\end{equation}
Here, the global advantage $\hat{A}_i^g$ is computed from the sample-level reward $R_i^g$ in Equation~\eqref{eq:sample_reward}. Early in training, a small $\alpha$ preserves the global quality constraint. The region signal is then gradually increased using a clipped cosine schedule:
\begin{equation}
\label{eq:alpha_warmup}
\alpha(s)=\alpha_{\min}
+\frac{\alpha_{\max}-\alpha_{\min}}{2}\left(1-\cos(\pi u(s))\right),
\end{equation}
where $u(s)=\operatorname{clip}\!\left((s-s_{\text{start}})/(s_{\text{end}}-s_{\text{start}}),0,1\right)$. Each region box is scaled to latent resolution and mapped to positions $M_{i,r}$; positions in $M_{i,r}$ use $\hat{A}_{i,r}^{\text{mix}}$, while uncovered positions use the background region. Overlapping region boxes and sparse semantic groups are handled by multi-region averaging and a global-advantage fallback, respectively; implementation details are provided in Appendix~\ref{sec:region_assignment_details}.

\subsection{Region-aligned policy objective}
\label{sec:region_ratio}
\label{sec:region_agg}
\label{sec:full_obj}

Region advantages describe editing-quality differences of semantic regions relative to the candidates in the same group. To preserve this correspondence during the update, the policy objective should also operate at the region level. If policy ratios and loss aggregation are still computed independently over latent positions, the region-level signal can be diluted by local noise along the diffusion trajectory and by the large area of the background. \method{} therefore introduces two region-aligned designs in the policy objective.

First, we aggregate position-level log-ratios within each region:
\begin{equation}
\label{eq:token_log_ratio}
\delta_{i,t,k}
=\log p_\theta(x_{t-1}^i\!\mid x_t^i,c)_k
-\log p_{\theta_{\text{old}}}(x_{t-1}^i\!\mid x_t^i,c)_k,
\end{equation}
\begin{equation}
\label{eq:region_ratio_def}
\bar{\delta}_{i,t,r}
=\frac{1}{|M_{i,r}|}\sum_{k\in M_{i,r}}\delta_{i,t,k},
\qquad
s_{i,t,r}=\exp(\bar{\delta}_{i,t,r}).
\end{equation}
All latent positions in the same region share the region ratio $s_{i,t,r}$, so the measure of policy change has the same granularity as the region advantage.

Second, to prevent whole-image averaging from weakening foreground signals, we average within each region and use power-weighted aggregation across regions:
\begin{equation}
\label{eq:token_surrogate}
\ell_{i,t,k}=
\max\!\left(
-\hat{A}_{i,k}^{\text{mix}}\,s_{i,t,r(k)},\;
-\hat{A}_{i,k}^{\text{mix}}\,\operatorname{clip}(s_{i,t,r(k)},1-\epsilon_-,1+\epsilon_+)
\right),
\end{equation}
where $r(k)$ is the region that contains position $k$. The policy loss is:
\begin{equation}
\label{eq:region_agg}
L_{\text{pg}}^{(t)}=
\frac{1}{G}\sum_{i=1}^{G}
\sum_{r\in\mathcal{R}_i}\tilde{w}_{i,r}
\frac{1}{|M_{i,r}|}
\sum_{k\in M_{i,r}}\ell_{i,t,k},
\end{equation}
where
\begin{equation}
\label{eq:region_weight}
\tilde{w}_{i,r}=
\frac{(|M_{i,r}|+\tau)^\beta}
{\sum_{r'\in\mathcal{R}_i}(|M_{i,r'}|+\tau)^\beta},
\quad \beta\in[0,1],\ \tau>0.
\end{equation}
When $\beta=1$, aggregation is area-weighted; when $\beta=0$, regions are weighted equally. We use $\beta=0.7$ and $\tau=10^{-2}$ to preserve area information while increasing foreground influence.

The final training objective is
\begin{equation}
\label{eq:unified_obj}
\mathcal{L}_{\text{\method{}}}(\theta)=
\frac{1}{|\mathcal{T}_s|}\sum_{t\in\mathcal{T}_s}\omega_t L_{\text{pg}}^{(t)}
\;+\;
\lambda_{\mathrm{kl}}\frac{1}{|\mathcal{T}_s|}\sum_{t\in\mathcal{T}_s}L_{\mathrm{kl}}^{(t)},
\end{equation}
where $\mathcal{T}_s$ is the sampled timestep set, $\omega_t$ follows Flow-GRPO, and $L_{\mathrm{kl}}^{(t)}$ regularizes against the reference policy. Unlike Flow-GRPO, \method{} keeps semantic-region granularity in advantages, ratios, and aggregation, preserving spatial attribution during updates.

The full training procedure, including region rewards, semantic advantage estimation, and the region-aligned policy objective, is summarized in Algorithm~\ref{alg:finegrpo} in Appendix~\ref{sec:training_algorithm}.

\section{Experiments}
\label{sec:experiments}

We evaluate the reward model, standard editing performance, multi-region editing ability, and component ablations.

\subsection{Experimental setup}
\label{sec:exp_setup}

We evaluate \method{} on \textbf{OmniGen2}~\citep{omnigen2_2025}, a Diffusion Transformer~\citep{peebles2023dit}, and \textbf{FLUX.2-klein-4B}~\citep{blackforestlabs2026flux2klein}, a Rectified Flow model~\citep{lipman2023flow,esser2024flux}. The main baseline is Flow-GRPO~\citep{flowgrpo2025}. Both methods use the same reward model, training data, base model, and sampling settings; Flow-GRPO uses the sample-level score in Equation~\eqref{eq:sample_reward} as the reward, while \method{} uses region scores to construct region advantages and the region-aligned policy objective.

Metrics include three benchmarks: GEdit-Bench~\citep{step1xedit2025}, ImgEdit-Bench~\citep{imgedit2025}, and MultiEditBench. RL post-training uses the 9-category editing subset of EditScore-RL-Data~\citep{editscore2025} as the source of editing instructions and source images. Training parameters and implementation details are provided in Appendix~\ref{sec:hyperparams}.

\subsection{\rewardmodel{} evaluation}
\label{sec:reward_model_eval}

We compare \rewardmodel{} with closed- and open-source evaluators on EditReward-Bench~\citep{editscore2025} and MMRB2~\citep{mmrb2_2025}. \rewardmodel{} is initialized from Qwen3-VL-8B-Instruct~\citep{qwen3vl2025}; the ``w/o Dense Anno.'' variant uses the same data source but keeps only the original global annotations, without the added region-level annotations. Table~\ref{tab:reward_model_bench} shows that dense region supervision improves the evaluator's ability to judge editing quality.

\begin{figure*}[!t]
\centering
\includegraphics[width=\textwidth]{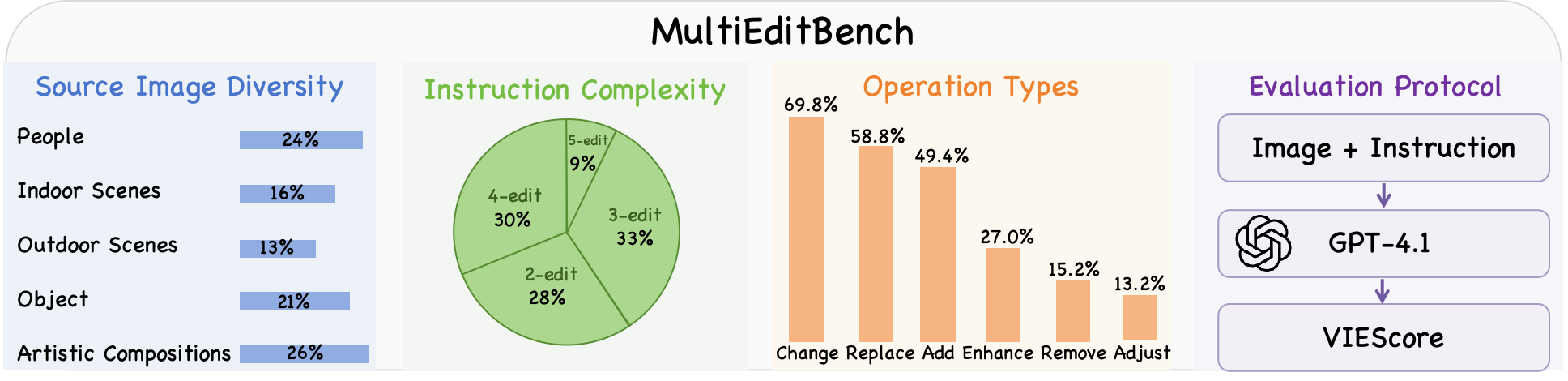}
\caption{\textbf{Composition of MultiEditBench.}}
\label{fig:multieditbench_overview}
\end{figure*}

\begin{table}[!t]
\centering
\caption{\textbf{\rewardmodel{} evaluation results.} PF, Cons., and Ovrl. denote prompt following, consistency, and overall, respectively.}
\label{tab:reward_model_bench}
\small
\setlength{\tabcolsep}{3pt}
\begin{tabular}{@{}llcccccc@{}}
\toprule
\multirow{2}{*}{Model} & \multirow{2}{*}{Type} & \multicolumn{3}{c}{EditReward-Bench} & \multicolumn{3}{c}{MMRB2} \\
\cmidrule(lr){3-5}\cmidrule(lr){6-8}
 & & PF & Cons. & Ovrl. & Single & Multi & Ovrl. \\
\midrule
\multicolumn{8}{l}{\textit{Closed-source models}} \\
GPT-4.1 & Closed & 0.673 & 0.602 & 0.705 & 0.547 & 0.478 & 0.535 \\
GPT-5 & Closed & 0.777 & 0.669 & 0.755 & 0.627 & 0.584 & 0.619 \\
Gemini-2.5-Pro & Closed & 0.703 & 0.560 & 0.722 & 0.545 & 0.483 & 0.534 \\
Gemini-3.0-Flash & Closed & 0.717 & 0.662 & \textbf{0.769} & 0.627 & 0.596 & \textbf{0.621} \\
\midrule
\multicolumn{8}{l}{\textit{Open-source generative evaluators}} \\
Qwen3-VL-8B & Gen. & 0.419 & 0.243 & 0.562 & 0.425 & 0.393 & 0.419 \\
EditScore-7B (Avg@4) & Gen. & 0.722 & 0.720 & 0.727 & -- & -- & -- \\
w/o Dense Anno. (Ours) & Gen. & 0.627 & 0.514 & 0.699 & 0.591 & 0.545 & 0.583 \\
\rowcolor{bestcolor} \rewardmodel{} (Ours) & Gen. & 0.648 & 0.621 & \textbf{0.760} & 0.629 & 0.551 & \textbf{0.615} \\
\bottomrule
\end{tabular}
\end{table}

\subsection{Main results}
\label{sec:main_results}

We further evaluate the effectiveness of \method{} as a region-level RL post-training method. Table~\ref{tab:standard_results} reports standard benchmark results on two base models. Across both OmniGen2 and FLUX.2-klein-4B, \method{} consistently improves over Flow-GRPO.

\begin{table}[t]
\centering
\caption{\textbf{Standard benchmark results.} Overall is the score after post-training, and $\Delta$ is the gain over the corresponding base model without RL.}
\label{tab:standard_results}
\small
\setlength{\tabcolsep}{2.5pt}
\begin{tabular}{llcccc}
\toprule
Base model & Method & GEdit Overall $\uparrow$ & $\Delta$ GEdit $\uparrow$ & ImgEdit Overall $\uparrow$ & $\Delta$ ImgEdit $\uparrow$ \\
\midrule
OmniGen2 & No RL & 6.42 & 0.00 & 3.44 & 0.00 \\
OmniGen2 & Flow-GRPO & 6.99 & +0.57 & 3.66 & +0.22 \\
\rowcolor{bestcolor} OmniGen2 & \method{} & \textbf{7.29} & \textbf{+0.87} & \textbf{3.74} & \textbf{+0.30} \\
\midrule
FLUX.2-klein-4B & No RL & 6.58 & 0.00 & 3.42 & 0.00 \\
FLUX.2-klein-4B & Flow-GRPO & 6.73 & +0.15 & 3.50 & +0.08 \\
\rowcolor{bestcolor} FLUX.2-klein-4B & \method{} & \textbf{6.80} & \textbf{+0.22} & \textbf{3.53} & \textbf{+0.11} \\
\bottomrule
\end{tabular}
\end{table}

These gains are consistent with the difference in optimization granularity. If region-level feedback is compressed into a sample-level score, local error sources are averaged during the update. In contrast, \method{} preserves region attribution in advantage estimation, policy ratios, and loss aggregation, allowing feedback to be converted into editing improvements more effectively. This trend appears on both Diffusion Transformer and Rectified Flow backbones, supporting the effectiveness of fine-grained credit assignment itself. Figure~\ref{fig:qualitative_maintext} provides qualitative examples of this behavior, with more results in Appendix~\ref{sec:qualitative_example}.

\begin{figure}[b]
\centering
\includegraphics[width=\linewidth]{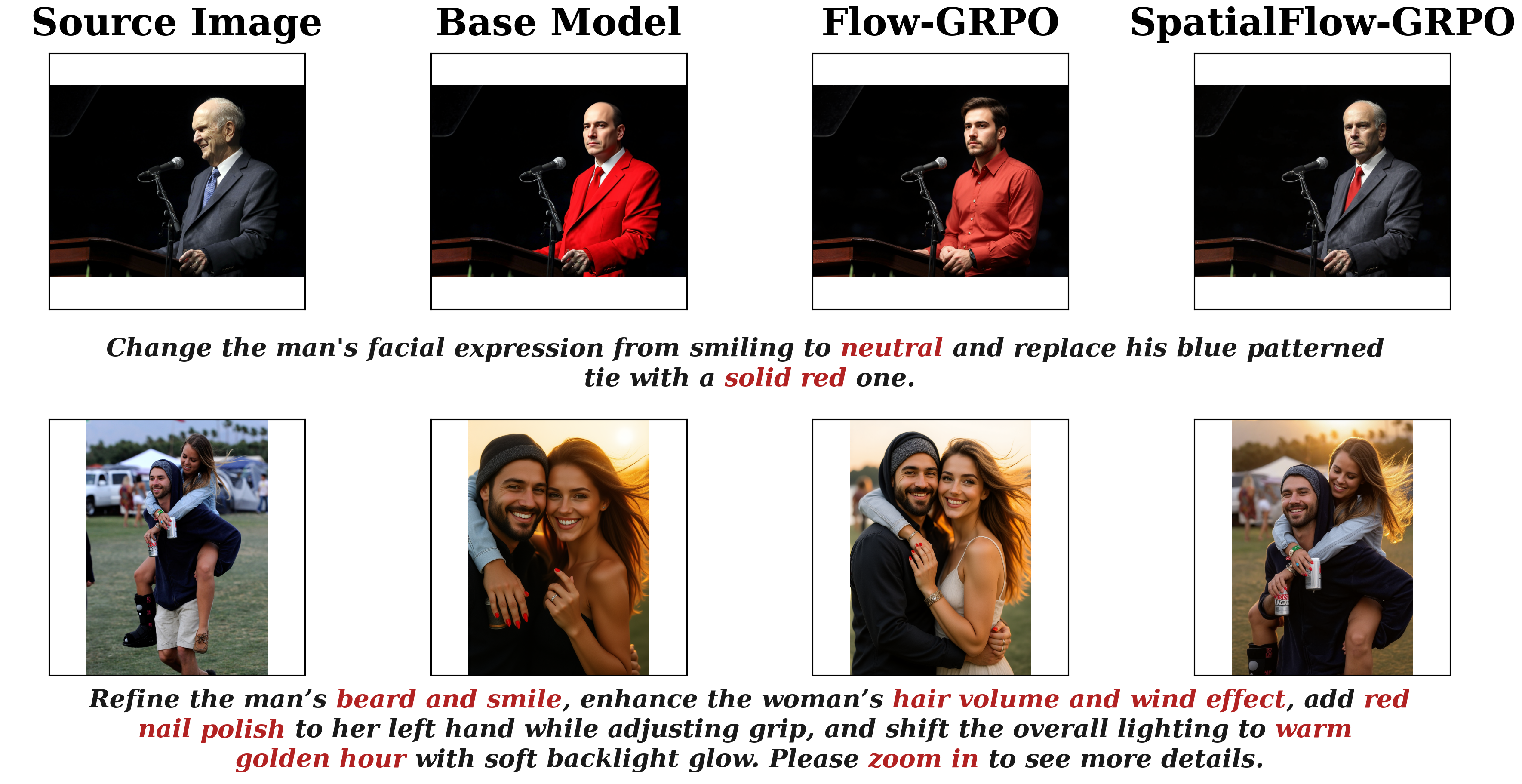}
\caption{\textbf{Qualitative examples of \method{} on OmniGen2.} Compared with Flow-GRPO, \method{} better preserves source identity and applies multiple requested local edits.}
\label{fig:qualitative_maintext}
\end{figure}

\subsection{MultiEditBench: multi-region editing diagnosis}
\label{sec:multieditbench}

MultiEditBench evaluates a model's ability to follow complex instructions that require multiple simultaneous edits. It contains 500 manually reviewed samples with 2--5 editing targets, which are grouped into 2-edit, 3-edit, 4-edit, and 5-edit subsets. We use GPT-4.1 with the VIEScore protocol~\citep{viescore2024} to judge semantic consistency (SC) and perceptual quality (PQ). Figure~\ref{fig:multieditbench_overview} summarizes the benchmark composition, and the construction process is described in Appendix~\ref{sec:multieditbench_details}.

Table~\ref{tab:multi_edit} compares \method{} with Flow-GRPO, the corresponding base models, and representative open-source editors. Across both OmniGen2 and FLUX.2-klein-4B, \method{} gives the strongest result among methods using the same backbone, suggesting that region-level credit assignment improves multi-target editing beyond standard whole-image reward optimization.

\begin{table}[t]
\centering
\caption{\textbf{Full MultiEditBench results.} MEB Score is reported by the number of simultaneous editing regions. Avg is the equal-weighted average over the four difficulty subsets. $\Delta$ Avg denotes the improvement over the corresponding base model and is omitted for standalone reference editors.}
\label{tab:multi_edit}
\small
\setlength{\tabcolsep}{2.2pt}
\begin{tabular}{@{}lcccc>{\columncolor{avgcolor}}cc@{}}
\toprule
Model & 2-edit & 3-edit & 4-edit & 5-edit & Avg & $\Delta$ Avg \\
 & ($n$=140) & ($n$=165) & ($n$=150) & ($n$=45) & \cellcolor{avgcolor} & \\
\midrule
Qwen-Image-Edit-2509 & 8.39 & 8.39 & 8.33 & 8.23 & 8.33 & -- \\
OmniGen2 + \method{} & 8.18 & 8.19 & 8.04 & 8.03 & 8.11 & \textbf{+1.70} \\
FLUX.2-klein-4B + \method{} & 8.03 & 8.07 & 8.14 & 7.92 & 8.04 & \textbf{+0.19} \\
FLUX.2-klein-4B + Flow-GRPO & 7.98 & 8.06 & 7.99 & 7.84 & 7.97 & +0.12 \\
FLUX.2-klein-4B Base & 7.84 & 7.85 & 7.99 & 7.73 & 7.85 & -- \\
OmniGen2 + Flow-GRPO & 7.52 & 7.42 & 7.40 & 7.72 & 7.52 & +1.11 \\
Step1X-Edit & 7.28 & 7.12 & 7.12 & 7.14 & 7.16 & -- \\
FLUX.1-Kontext & 6.80 & 7.04 & 6.80 & 7.66 & 7.07 & -- \\
BAGEL-7B-MoT & 6.67 & 6.74 & 6.50 & 6.59 & 6.62 & -- \\
OmniGen2 Base & 6.80 & 6.43 & 6.32 & 6.09 & 6.41 & -- \\
\bottomrule
\end{tabular}
\end{table}

\subsection{Component and parameter ablations}
\label{sec:ablation}

The ablation study analyzes \method{} from two aspects: parameter sensitivity and component contribution.

\begin{table}[htbp]
\centering
\caption{\textbf{$\beta$--$\alpha$ parameter scan} (OmniGen2, \rewardmodel{} reward model). Overall is the score, and $\Delta$ is the gain over the OmniGen2 base model (GEdit 6.42, ImgEdit 3.44). ``w'' indicates cosine warmup.}
\label{tab:ablation_full}
\small
\setlength{\tabcolsep}{4pt}
\begin{tabular}{cccccc}
\toprule
$\beta$ & $\alpha$ & \makecell{GEdit\\Overall $\uparrow$} & \makecell{$\Delta$\\GEdit $\uparrow$} & \makecell{ImgEdit\\Overall $\uparrow$} & \makecell{$\Delta$\\ImgEdit $\uparrow$} \\
\midrule
\multicolumn{6}{l}{\textit{$\beta=1$ (token-weighted, equivalent to global averaging)}} \\
1 & 1 (pure region advantage) & 6.92 & +0.50 & 3.63 & +0.19 \\
1 & 0.5 & 7.04 & +0.62 & 3.68 & +0.24 \\
1 & 0.7 & 6.98 & +0.56 & 3.64 & +0.20 \\
1 & 0.3 & 7.02 & +0.60 & 3.68 & +0.24 \\
1 & 0.5w & 7.07 & +0.65 & 3.67 & +0.23 \\
\midrule
\multicolumn{6}{l}{\textit{$\beta=0$ (equal region weighting)}} \\
0 & 1 & 6.94 & +0.52 & 3.63 & +0.19 \\
0 & 0.5 & 7.12 & +0.70 & 3.70 & +0.26 \\
0 & 0.5w & 7.17 & +0.75 & 3.69 & +0.25 \\
0 & 0.7w & 7.01 & +0.59 & 3.66 & +0.22 \\
\midrule
\multicolumn{6}{l}{\textit{$\beta=0.7$ (power-weighted trade-off)}} \\
\rowcolor{bestcolor} 0.7 & 0.5w & \textbf{7.29} & \textbf{+0.87} & \textbf{3.74} & \textbf{+0.30} \\
0.7 & 0.5 & 7.21 & +0.79 & 3.72 & +0.28 \\
0.7 & 0.3w & 7.05 & +0.63 & 3.66 & +0.22 \\
\midrule
\multicolumn{6}{l}{\textit{Other $\beta$ values}} \\
0.5 & 0.5w & 7.10 & +0.68 & 3.68 & +0.24 \\
0.9 & 0.5w & 7.09 & +0.67 & 3.68 & +0.24 \\
\bottomrule
\end{tabular}
\end{table}

Table~\ref{tab:ablation_full} scans the region-mixing coefficient $\alpha$ and the region aggregation power $\beta$. The coefficient $\alpha$ controls the trade-off between local region advantages and the global advantage. When the region signal is too strong or introduced too early, updates can be dominated by local scores, weakening global quality constraints such as composition, identity preservation, and naturalness. The parameter $\beta$ controls how sensitive region aggregation is to area: aggregation close to area weighting can still dilute foreground signals by large background regions, while equal region weighting can over-amplify high-variance small regions. These results suggest that region feedback should be introduced gradually and under global constraints.

\begin{table}[t]
\centering
\caption{\textbf{Progressive ablation.} Starting from a Flow-GRPO baseline using the sample-level score from \rewardmodel{}, we progressively add region advantages, global--region mixed advantages, $\alpha$ warmup, and region-balanced aggregation. The corresponding $\beta$/$\alpha$ settings are also shown.}
\label{tab:incremental}
\small
\setlength{\tabcolsep}{2.4pt}
\begin{tabular}{@{}lccccccccc@{}}
\toprule
\makecell{Variant} & \makecell{Reg.\\adv.} & \makecell{Mixed\\adv.} & \makecell{$\alpha$\\warm.} & \makecell{Reg.\\ratio} & \makecell{Bal.\\agg.} & $\beta$ & $\alpha$ & \makecell{$\Delta$\\GEdit} & \makecell{$\Delta$\\ImgEdit} \\
\midrule
V1 (Flow-GRPO) & \xmark & \xmark & \xmark & \xmark & \xmark & -- & -- & 0.57 & 0.22 \\
V2 (+Reg. adv.) & \cmark & \xmark & \xmark & \cmark & \xmark & 1 & 1 & 0.50 & 0.19 \\
V3 (+Mix.) & \cmark & \cmark & \xmark & \cmark & \xmark & 0 & 0.5 & 0.70 & 0.26 \\
V4 (+Warmup) & \cmark & \cmark & \cmark & \cmark & \xmark & 0 & 0.5w & 0.75 & 0.25 \\
\rowcolor{bestcolor} V5 (Ours)
& \cmark & \cmark & \cmark & \cmark & \cmark & 0.7 & 0.5w & \textbf{0.87} & \textbf{0.30} \\
\bottomrule
\end{tabular}
\end{table}

Table~\ref{tab:incremental} further shows that adding region advantages alone does not stably outperform Flow-GRPO, indicating that fine-grained rewards provide more localized supervision but can also amplify local noise without proper constraints. Global--region mixing provides an image-level quality anchor, $\alpha$ warmup prevents early training from relying too heavily on unstable local signals, and region-balanced aggregation reduces the dilution of foreground edits by background area. Overall, the benefit of \method{} comes from the cooperation of region rewards, global quality constraints, and the region-aligned policy objective, rather than from simply replacing sample-level rewards with region rewards.

\FloatBarrier

\section{Conclusion and discussion}
\label{sec:conclusion}

We presented \method{} for spatial credit assignment in image editing. It treats semantic regions as fine-grained optimization units and uses \rewardmodel{} and \rewarddata{} to provide region-level feedback. On OmniGen2 and FLUX.2-klein-4B, \method{} consistently outperforms Flow-GRPO, with larger gains on MultiEditBench multi-region edits, showing that semantically grouped rewards better capture local editing quality. The method still relies on region annotations and reward-model reliability; future work can study more flexible region representations and larger-scale human evaluation.

\newpage
\bibliographystyle{plainnat}
\bibliography{refs}

\newpage
\appendix
\section{Additional details}

\subsection{Hyperparameters and training details}
\label{sec:hyperparams}

Table~\ref{tab:hyperparams} lists the actual RL post-training configuration used for the main OmniGen2 experiments with \method{}. All RL post-training experiments use the 9-category editing subset of EditScore-RL-Data~\citep{editscore2025}.

\begin{table}[h]
\centering
\caption{\textbf{Main hyperparameters.}}
\label{tab:hyperparams}
\small
\setlength{\tabcolsep}{4pt}
\begin{tabular}{ll}
\toprule
Configuration & Value \\
\midrule
\multicolumn{2}{l}{\textit{Data and sampling}} \\
RL training data & EditScore-RL-Data 9-category editing subset \\
Maximum text tokens & 888 \\
Output resolution cap & $512\times512$ \\
Unique instructions per round & 48 \\
Samples per instruction $G$ & 12 \\
Inference timesteps & 20 \\
Maximum generation length & 1024 \\
Text/image guidance scale & 4 / 2 \\
CFG training interval & $[0.0, 0.6]$ \\
Training timestep ratio & 0.6 \\
\midrule
\multicolumn{2}{l}{\textit{Optimization and finetuning}} \\
Global batch size & 576 \\
Training batch size & 36 \\
Gradient accumulation steps & 2 \\
Total training steps & 2000 \\
Learning rate & $4\times10^{-4}$ \\
Learning-rate schedule & timm\_constant\_with\_warmup \\
Optimizer & AdamW ($\beta_1=0.9,\beta_2=0.95$) \\
Weight decay / $\epsilon$ & 0.01 / $10^{-8}$ \\
Maximum gradient norm & 1 \\
Mixed precision & bf16 \\
Gradient checkpointing & Enabled \\
Finetuning strategy & LoRA~\citep{hu2022lora} ($r$=32, $\alpha$=64, dropout=0) \\
\midrule
\multicolumn{2}{l}{\textit{\method{} specific}} \\
Update steps after sampling & 2 \\
Forward batch size & 9 \\
PPO clip range & $[10^{-4}, 5\times10^{-4}]$ \\
Advantage clipping threshold & 5 \\
KL weight $\lambda_{\mathrm{kl}}$ & 0.04 \\
Sampling noise coefficient & 0.7 \\
Mixing coefficient $\alpha_{\max}$ & 0.5 \\
Mixing coefficient $\alpha_{\min}$ & 0.0 \\
Warmup start/end steps $s_{\text{start}}$ / $s_{\text{end}}$ & 0 / 500 \\
Region power $\beta$ & 0.7 \\
Smoothing term $\tau$ & $10^{-2}$ \\
Policy-loss reweighting & Enabled \\
\bottomrule
\end{tabular}
\end{table}

\subsection{Training algorithm}
\label{sec:training_algorithm}

\begin{algorithm}[H]
\caption{\method{} training}
\label{alg:finegrpo}
\begin{algorithmic}[1]
\Require Policy model $\theta$, reference model $\theta_{\text{ref}}$, region reward model $\mathcal{R}$, mixing parameter $\alpha(s)$, power parameter $\beta$
\For{each training round}
  \State Sample $G$ edited images for each editing instruction $p$ and record old-policy log probabilities
  \State Use $\mathcal{R}$ to evaluate structured region scores $\{R_{i,r}\}$
  \State Compute region advantages $\hat{A}_{i,r}$ by grouping over $(p,l)$ (Equation~\ref{eq:region_adv})
  \State Compute the global advantage $\hat{A}_i^g$ and mix it into $\hat{A}_{i,r}^{\text{mix}}$ (Equation~\ref{eq:hybrid_adv})
  \State Map $\hat{A}_{i,r}^{\text{mix}}$ to latent positions to form a fine-grained advantage map
  \For{each update step}
    \State Compute current-policy position-level log probabilities
    \State Compute the region-consistent ratio $s_{i,t,r}$ (Equation~\ref{eq:region_ratio_def})
    \State Compute the power-weighted region policy loss $L_{\text{pg}}^{(t)}$ (Equation~\ref{eq:region_agg})
    \State Compute the KL regularization term $L_{\mathrm{kl}}^{(t)}$
    \State Update $\theta$ using Equation~\eqref{eq:unified_obj}
  \EndFor
\EndFor
\end{algorithmic}
\end{algorithm}

\subsection{Region assignment and sparse semantic groups}
\label{sec:region_assignment_details}

\paragraph{Overlapping region boxes.}
\rewardmodel{} predicts semantic bounding boxes, so different edited regions may overlap after being mapped to the latent resolution. For a latent position $k$, let $\mathcal{C}_i(k)=\{r\in\mathcal{R}_i^{\mathrm{fg}}\mid k\in M_{i,r}\}$ denote the foreground regions that cover it. If $\mathcal{C}_i(k)$ is non-empty, the position-level mixed advantage is the average of $\hat{A}_{i,r}^{\mathrm{mix}}$ over $r\in\mathcal{C}_i(k)$, and the corresponding region-ratio term is averaged over the same covering regions. If $\mathcal{C}_i(k)$ is empty, the position is assigned to the background region. This gives each latent position a unique and well-defined training signal while allowing overlapping positions to receive credit from all relevant semantic regions.

\paragraph{Sparse semantic labels.}
Equation~\eqref{eq:region_adv} is used only when a semantic group contains at least two valid region rewards and has non-degenerate variance. If a label appears only once in the sampled group for an instruction, or if the within-group variance is too small, the group does not provide a reliable relative comparison. In this case, we do not compute an unstable region z-score; instead, the region advantage falls back to the corresponding sample-level global advantage $\hat{A}_i^g$. Thus the update reduces to global credit assignment when region-level comparison is underdetermined, and uses semantic-region relative advantages only when they are well defined.

\subsection{\rewardmodel{} and \rewarddata{} details}
\label{sec:reward_details}

\subsubsection{\rewarddata{} construction}
\label{sec:reward_data_details}

\begin{figure}[h]
\centering
\includegraphics[width=\linewidth]{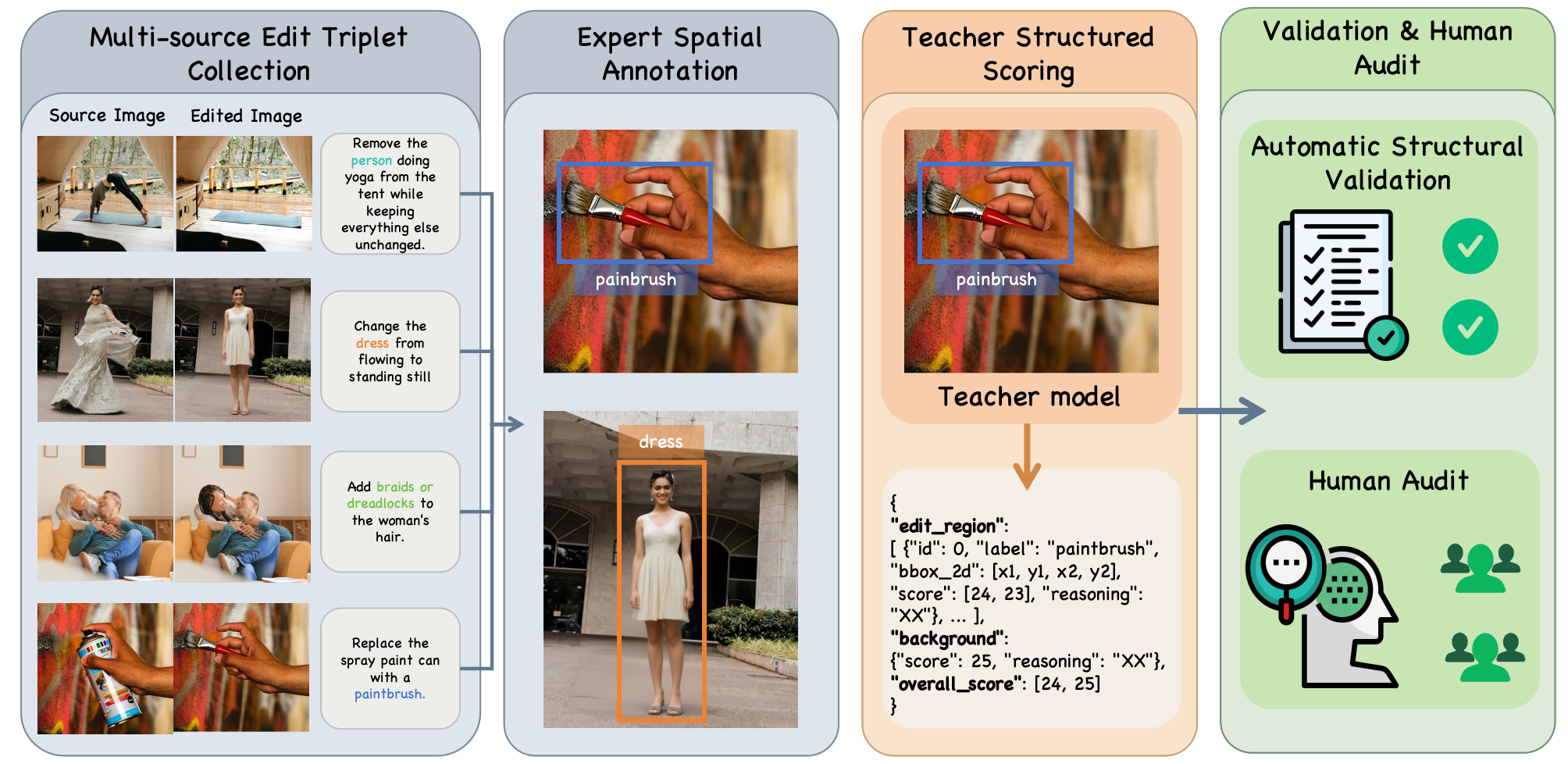}
\caption{\textbf{\rewarddata{} construction pipeline.} Multi-source edit triplets are annotated with expert region boxes and labels, scored by a multimodal teacher, and filtered through automatic validation and human audit to produce data for training a region-aware reward model.}
\label{fig:data_pipeline}
\end{figure}

The region-aware reward model used in our experiments is denoted as \rewardmodel{}. To train this model, we build \rewarddata{} from multi-source edit triplets. As shown in Figure~\ref{fig:data_pipeline}, the pipeline collects a source image, an edited image, and an instruction, asks expert annotators to mark normalized bounding boxes and semantic labels for edited regions, uses Gemini-3-Pro~\citep{gemini2025} to produce per-region success/preserve scores, background preservation scores, overall scores, and reasoning, and then applies automatic structural validation followed by human audit. This pipeline explicitly binds reward scores to semantic regions, providing input signals for region advantage estimation and local policy updates in \method{}. \rewarddata{} will be released to support reproduction of \rewardmodel{} training and research on region rewards.

\subsubsection{\rewardmodel{} training setup}
\label{sec:reward_model_training}

\rewardmodel{} is initialized from Qwen3-VL-8B-Instruct~\citep{qwen3vl2025} and trained on \rewarddata{} with supervised finetuning. Table~\ref{tab:reward_model_training} gives the main training settings.

\begin{table}[h]
\centering
\caption{\textbf{\rewardmodel{} training hyperparameters.}}
\label{tab:reward_model_training}
\small
\begin{tabular}{ll}
\toprule
Configuration & Value \\
\midrule
Base model & Qwen3-VL-8B-Instruct \\
Training stage & SFT \\
Finetuning method & LoRA ($r$=32, $\alpha$=64, dropout=0.05) \\
LoRA target modules & all \\
Training data & dense\_grpo\_coldstart\_v1 \\
Input template & qwen3\_vl \\
Maximum sequence length & 8192 \\
Maximum image/video pixels & 262144 / 16384 \\
Per-device batch size & 4 \\
Gradient accumulation steps & 4 \\
Learning rate & 1e-4 \\
Training epochs & 10 \\
Learning-rate schedule & cosine, warmup ratio 0.1 \\
Optimizer & AdamW \\
Training precision & bfloat16 \\
Preprocessing/loading workers & 192 / 48 \\
\bottomrule
\end{tabular}
\end{table}

\subsubsection{Scoring prompt for the region-aware reward model}
\label{sec:reward_prompt}

The following prompt is used to label \rewarddata{} with the Gemini-3-Pro teacher model. It instructs the model to score each editing region in a structured and multi-dimensional way with reasoning.

\begin{promptblock}
You are a professional digital artist. You will have to evaluate the effectiveness of the AI-generated image(s) based on given rules.\\[2pt]
All the input images are AI-generated.\\[4pt]
\textbf{OUTPUT FORMAT:}\\
\{\\
\quad "edit\_region": [\\
\quad\quad \{"id": 0, "label": "region label",\\
\quad\quad\quad "bbox\_2d": [x1, y1, x2, y2],\\
\quad\quad\quad "score": [score\_success, score\_preserve],\\
\quad\quad\quad "reasoning": "brief reason"\},\\
\quad\quad ...\\
\quad ],\\
\quad "background": \{"score": value, "reasoning": "..."\},\\
\quad "overall\_score": [overall\_success, overall\_preserve]\\
\}\\[4pt]
\textbf{RULES:}\\
Two images will be provided: original and edited version.\\
The objective is to evaluate how successfully the editing instruction has been executed.\\
You will be provided with pre-identified editing regions (bounding boxes with labels). Score each region separately.\\[4pt]
\textbf{SCORING (per region, 0--25):}\\
1) score\_success: how well the edit follows the instruction (0=no change, 25=perfect).\\
2) score\_preserve: degree of preservation within the region (0=completely different, 25=minimal effective edit).\\[4pt]
\textbf{BACKGROUND (0--25):}\\
Rate how well non-edited areas are preserved. Penalize unexpected edits, layout changes, artifacts outside editing regions.\\[4pt]
\textbf{OVERALL (0--25):}\\
Overall success score and overall overediting score.
\end{promptblock}

The prompt has three key design elements. First, \textbf{region-decoupled scoring} evaluates success and preserve scores independently for each edited region, while keeping whole-image scores as auxiliary information. Second, \textbf{spatial anchoring} binds scores to pre-annotated bounding boxes so that the scorer attends to the correct spatial location. Third, \textbf{reasoning} asks the scorer to provide a short explanation for each region, improving interpretability and consistency.

\subsubsection{Example output from the region reward model}
\label{sec:reward_example}

The following is a structured output example from the region-aware reward model for a multi-region editing instruction. The instruction is: ``Change the man's shirt to red and add a tree in the background.'' \rewardmodel{} first outputs semantic-consistency scores:

\begin{jsonbox}
{
  "edit_region": [
    {
      "id": 0, "label": "person's shirt",
      "bbox_2d": [120, 80, 420, 360],
      "score": [22, 20],
      "reasoning": "Shirt color successfully changed to 
        red; slight texture distortion at collar."
    },
    {
      "id": 1, "label": "tree",
      "bbox_2d": [520, 100, 860, 700],
      "score": [18, 15],
      "reasoning": "Tree added but appears slightly 
        unnatural; some blending artifacts with sky."
    }
  ],
  "background": {
    "score": 21,
    "reasoning": "Background well preserved; minor 
      color shift near added tree boundary."
  },
  "overall_score": [20, 18]
}
\end{jsonbox}

At the same time, \rewardmodel{} outputs perceptual-quality scores:
\begin{jsonbox}
{
  "reasoning": "The edited image is mostly natural, 
    with mild artifacts around the inserted tree.",
  "score": [22, 20]
}
\end{jsonbox}

The above output is converted into scalar region rewards by Equation~\eqref{eq:region_reward}. In this example, the \texttt{shirt} region receives a high reward because the edit succeeds and is well preserved. The \texttt{tree} region receives a lower reward because artifacts appear. The background region receives a moderate reward. Differentiated region rewards provide the basis for local credit assignment in \method{}.

\subsubsection{Training data statistics}
\label{sec:data_stats}

Table~\ref{tab:data_stats} reports statistics of \rewarddata{}. The data is filtered through multiple rounds of quality control, resulting in a final retention rate of 98.75\%.

\begin{table}[h]
\centering
\caption{\textbf{\rewarddata{} statistics.}}
\label{tab:data_stats}
\small
\begin{tabular}{ll}
\toprule
Statistic & Value \\
\midrule
Total samples & 14,276 \\
Average edited regions per sample & 2.32 \\
Maximum edited regions per sample & 7 \\
Single-region samples & 24.3\% \\
Multi-region samples & 75.7\% \\
\midrule
Edit type distribution & \\
\quad Attribute change (color/material/texture) & 38.2\% \\
\quad Object operation (add/replace/remove) & 33.5\% \\
\quad Background modification & 18.7\% \\
\quad Style/global transformation & 9.6\% \\
\midrule
Quality control & \\
\quad Initially collected samples & 16,465 \\
\quad Structured-output pass rate & 99.68\% \\
\quad Final retention rate & 98.75\% \\
\bottomrule
\end{tabular}
\end{table}

\subsection{MultiEditBench details}
\label{sec:multieditbench_details}

\subsubsection{Data construction and statistics}

MultiEditBench is designed to diagnose model performance in multi-region parallel editing scenarios. Source images come from several high-quality public sources and cover visual types such as people, indoor and outdoor scenes, object close-ups, and artistic compositions. The source distribution is shown in Table~\ref{tab:multieditbench_sources}. For each source image, we first use a VLM to identify editable elements. An LLM then generates compound instructions containing 2--5 edits. Human review removes ambiguous, contradictory, or overly simple samples.

\begin{table}[h]
\centering
\caption{\textbf{Source images in MultiEditBench.}}
\label{tab:multieditbench_sources}
\small
\begin{tabular}{lcl}
\toprule
Source & Ratio & Description \\
\midrule
LAION-Aesthetics & 42.4\% & Natural images with high aesthetic scores \\
CC8M & 42.0\% & Diverse web images \\
Pexels & 7.8\% & Professional photography images \\
Others & 7.8\% & Other high-quality sources \\
\bottomrule
\end{tabular}
\end{table}

Editing operations include attribute modification, object replacement, element addition, enhancement, removal, and position adjustment. Since a compound instruction usually contains multiple operation types, the frequencies in Table~\ref{tab:multieditbench_ops} are not mutually exclusive.

\begin{table}[h]
\centering
\caption{\textbf{Editing operation types in MultiEditBench.}}
\label{tab:multieditbench_ops}
\small
\begin{tabular}{lcl}
\toprule
Operation type & Frequency & Example \\
\midrule
Change (attribute modification) & 69.8\% & Change color, material, or expression \\
Replace (object replacement) & 58.8\% & Replace an object with another object \\
Add (element addition) & 49.4\% & Add a new object or decoration \\
Enhance (enhancement) & 27.0\% & Enhance lighting, makeup, or details \\
Remove (removal) & 15.2\% & Remove a specified element \\
Adjust (adjustment) & 13.2\% & Adjust pose, angle, or position \\
\bottomrule
\end{tabular}
\end{table}

\begin{table}[h]
\centering
\caption{\textbf{Difficulty distribution of MultiEditBench.} Samples are grouped by the number of simultaneous editing operations.}
\label{tab:multieditbench_stats}
\small
\begin{tabular}{lcccc}
\toprule
Type & Meaning & Count & Ratio & Avg. instruction length \\
\midrule
2-edit & 2 simultaneous edits & 140 & 28.0\% & 131 chars \\
3-edit & 3 simultaneous edits & 165 & 33.0\% & 169 chars \\
4-edit & 4 simultaneous edits & 150 & 30.0\% & 198 chars \\
5-edit & 5 simultaneous edits & 45 & 9.0\% & 256 chars \\
\midrule
Total & -- & 500 & 100\% & -- \\
\bottomrule
\end{tabular}
\end{table}

\subsubsection{MultiEditBench evaluation prompts}
\label{sec:eval_prompt}

MultiEditBench follows the VIEScore~\citep{viescore2024} framework and uses GPT-4.1~\citep{openai2025gpt41} as the evaluation backbone. Each sample is evaluated separately for semantic consistency (SC) and perceptual quality (PQ), and the final metric is $\text{MEB Score}=\sqrt{\text{SC}\times\text{PQ}}$. Scores are averaged within each task type. The final Avg is the equal-weighted average over the 2-edit, 3-edit, 4-edit, and 5-edit difficulty subsets.

\paragraph{Semantics Score (SC).}
SC takes the source image, edited image, and editing instruction as input. It outputs editing success and over-editing control scores, and their mean is used as SC.

\begin{promptblock}
You are a professional digital artist. You will evaluate the effectiveness of the AI-generated image(s).\\[4pt]
\textbf{RULES:}\\
Two images will be provided: the first being the original AI-generated image and the second being an edited version.\\
Evaluate how successfully the editing instruction has been executed in the second image.\\[4pt]
\textbf{SCORING (0--10):}\\
score1: editing success, where 0 means the edit does not follow the instruction and 10 means perfect instruction following.\\
score2: degree of overediting, where 0 means the edited image is completely different from the original and 10 means a minimal yet effective edit.\\
Output: \{"score": [score1, score2], "reasoning": "..."\}\\[4pt]
Editing instruction: \{instruction\}
\end{promptblock}

\paragraph{Quality Score (PQ).}
PQ takes only the edited image as input. It outputs naturalness and artifact-control scores, and their mean is used as PQ.

\begin{promptblock}
The image is an AI-generated image. Evaluate the generation quality.\\[4pt]
\textbf{SCORING (0--10):}\\
naturalness: 0 means the image does not look natural; 10 means it looks natural.\\
artifacts: 0 means severe distortion, watermark, blur, unusual body parts, or disharmonized subjects; 10 means no artifacts.\\
Output: \{"score": [naturalness, artifacts], "reasoning": "..."\}
\end{promptblock}

\subsection{Notation}
\label{sec:notation}

Table~\ref{tab:notation} summarizes the main notation used in this paper.

\begin{table}[h]
\centering
\caption{\textbf{Main notation.}}
\label{tab:notation}
\small
\begin{tabular}{cl}
\toprule
\textbf{Symbol} & \textbf{Meaning} \\
\midrule
$G$ & Number of samples per editing instruction \\
$T$ & Total number of diffusion timesteps \\
$c$ & Editing instruction (condition) \\
$R_{i,r}$ & Region reward of region $r$ in sample $i$ \\
$\hat{A}_{i,r}$ & Region advantage of region $r$ in sample $i$ \\
$\hat{A}_{i,r}^{\text{mix}}$ & Global--region mixed advantage \\
$\alpha(s)$ & Region mixing coefficient as a function of training step $s$ \\
$\beta$ & Region power-weighting parameter \\
$\tau$ & Smoothing term for power weighting \\
$M_{i,r}$ & Latent-position set corresponding to region $r$ \\
$s_{i,t,r}$ & Region-consistent ratio \\
$\mathcal{S}_{p,l}$ & Semantic group $(\text{instruction}=p, \text{label}=l)$ \\
$\epsilon_+, \epsilon_-$ & PPO clipping bounds \\
$\lambda_{\mathrm{kl}}$ & KL regularization weight \\
$\omega_t$ & Timestep reweighting coefficient \\
\bottomrule
\end{tabular}
\end{table}

\subsection{Statistical evaluation and compute resources}
\label{sec:stat_compute}

For the main RL comparisons, we repeat training with different random seeds.

The main RL post-training experiments were run on 16 NVIDIA H800 GPUs for about 40 hours, corresponding to roughly 640 GPU-hours for a main run. The same hardware class was used for Flow-GRPO baselines and \method{} variants. Compute for ablations scales approximately linearly with the number of training variants because each variant uses the same sampling and update schedule.

\subsection{Responsible research, assets, and LLM usage}
\label{sec:responsible_assets}

This work improves image editing models and may have both positive and negative societal impacts. Positive uses include controllable content creation, image restoration, and editing assistance; negative uses include misleading edits, identity manipulation, and disinformation. Released data and models will include usage restrictions, safety notes, and filtering for unsafe or privacy-sensitive content.

All existing models, datasets, and benchmarks are cited in the main text or bibliography. We follow their licenses and terms of use, and will provide the code in the supplementary material.

Human annotation is used to mark editing regions, assign semantic labels, and review benchmark samples. Annotators receive written instructions and examples, are compensated according to local requirements, and the task does not collect private information or involve interventions on human subjects.

LLMs and VLMs are used in the data and evaluation pipeline: Gemini-3-Pro produces structured region scores for \rewarddata{}, VLM/LLM tools identify editable elements and generate candidate MultiEditBench instructions before human review, and GPT-4.1 serves as the VIEScore-style evaluation backbone. These uses are described in Appendix~\ref{sec:reward_data_details} and Appendix~\ref{sec:eval_prompt}.

\clearpage
\subsection{Additional OmniGen2 qualitative comparisons}
\label{sec:qualitative_example}

\begin{figure}[H]
\centering
\makebox[\textwidth][c]{\includegraphics[height=0.86\textheight, keepaspectratio, page=1]{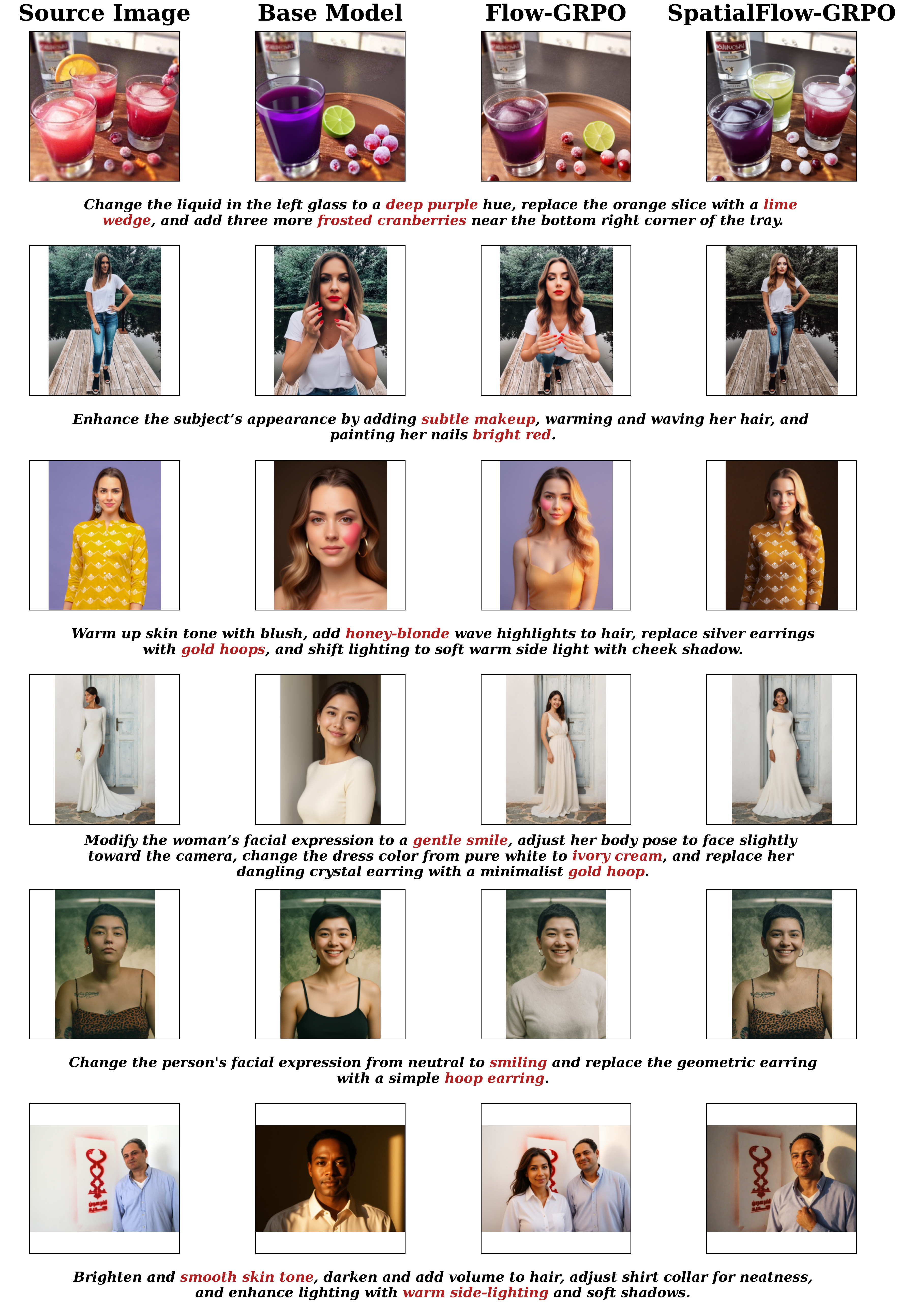}}
\caption{\textbf{Additional OmniGen2 qualitative comparisons} (1/3). Each row compares the base model, Flow-GRPO, and \method{} under multi-target editing instructions.}
\label{fig:qualitative_omnigen2}
\end{figure}

\begin{figure}[H]
\centering
\makebox[\textwidth][c]{\includegraphics[height=0.90\textheight, keepaspectratio, page=2]{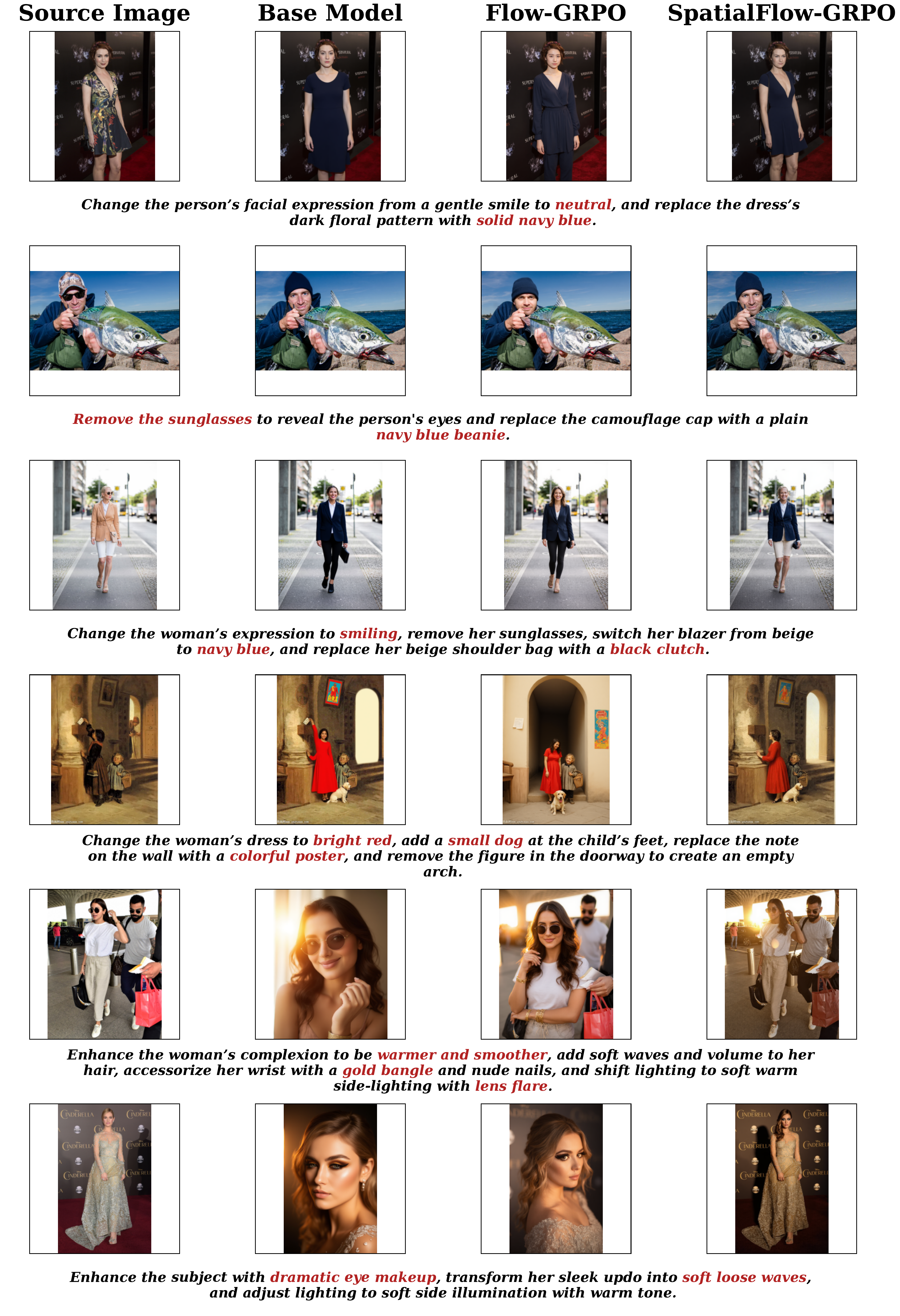}}
\caption{\textbf{Additional OmniGen2 qualitative comparisons} (2/3).}
\end{figure}

\begin{figure}[H]
\centering
\makebox[\textwidth][c]{\includegraphics[height=0.90\textheight, keepaspectratio, page=3]{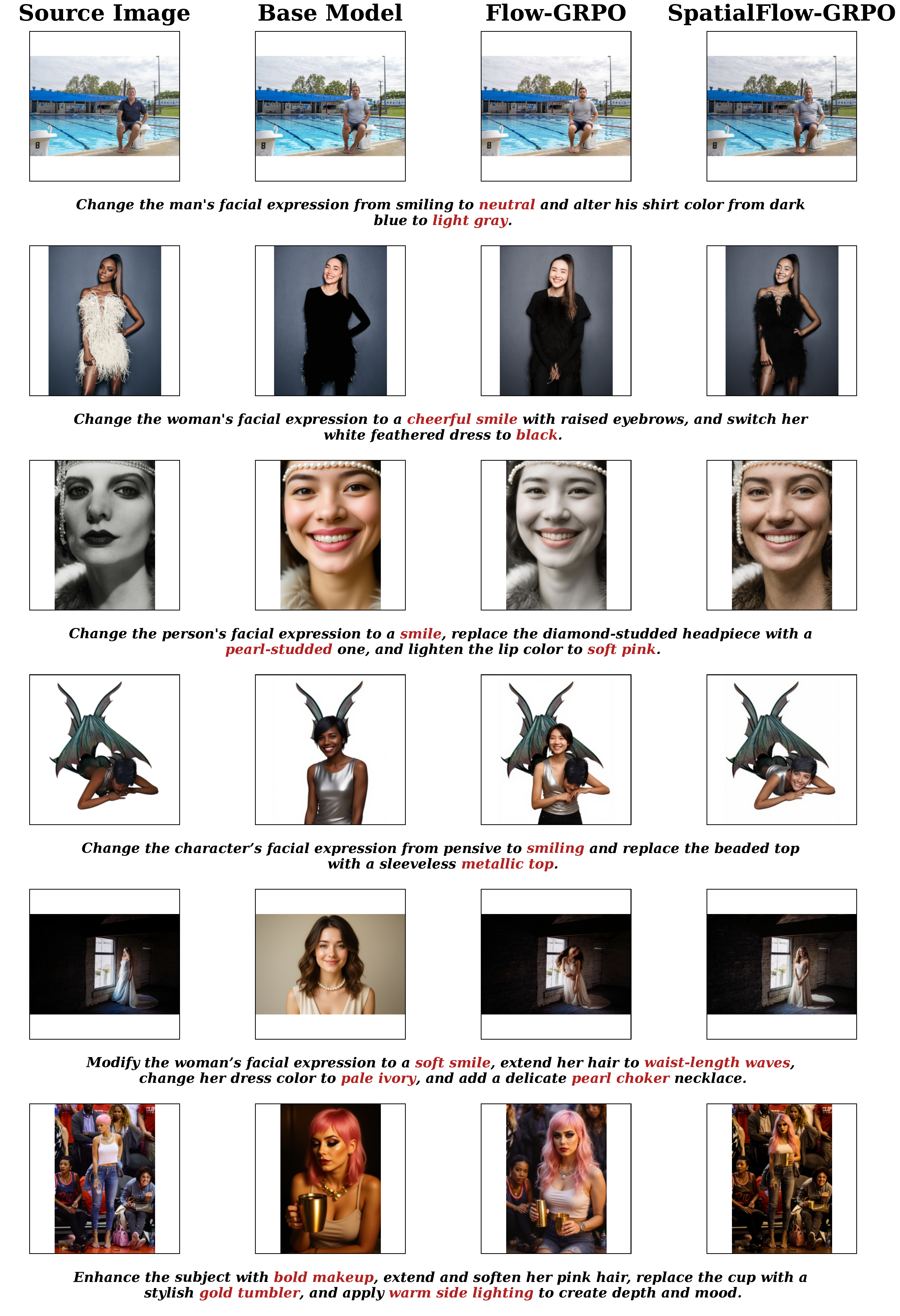}}
\caption{\textbf{Additional OmniGen2 qualitative comparisons} (3/3).}
\end{figure}

\end{document}